\documentclass[runningheads]{llncs}
\usepackage{dirtytalk}
\usepackage{graphicx}
\begin{document}
\title{Multi-script Handwritten Digit Recognition Using Multi-task Learning}
%
%
\author{Mesay Samuel Gondere\inst{1} \and
Lars Schmidt-Thieme\inst{2} \and
Durga Prasad Sharma\inst{1} \and
Randolf Scholz\inst{2}}
\authorrunning{M. Samuel et al.}
%
\institute{Arba Minch University, Faculty of Computing and Software Engineering, Ethiopia
\email{\{mesay.samuel,sharma.dp\}@amu.edu.et} \and
Information Systems and Machine Learning Lab, 31141 Hildesheim, Germany
\email{\{schmidt-thieme,scholz\}@ismll.uni-hildesheim.de}}

\maketitle              
\begin{abstract}
Handwritten digit recognition is one of the extensively studied area in machine learning. Apart from the wider research on handwritten digit recognition on MNIST dataset, there are many other research works on various script recognition. However, it is not very common for multi-script digit recognition which encourage the development of robust and multipurpose systems. Additionally working on multi-script digit recognition enables multi-task learning, considering the script classification as a related task for instance. It is evident that multi-task learning improves model performance through inductive transfer using the information contained in related tasks. Therefore, in this study multi-script handwritten digit recognition using multi-task learning will be investigated. As a specific case of demonstrating the solution to the problem, Amharic handwritten character recognition will also be experimented. The handwritten digits of three scripts including Latin, Arabic and Kannada are studied to show that multi-task models with reformulation of the individual tasks have shown promising results. In this study a novel way of using the individual tasks predictions was proposed to help classification performance and regularize the different loss for the purpose of the main task. This finding has outperformed the baseline and the conventional multi-task learning models. More importantly, it avoided the need for weighting the different losses of the tasks, which is one of the challenges in multi-task learning.

\end{abstract}
\section{Introduction}
Handwritten digit recognition is commonly known to be the \say{Hello World} of machine learning. Accordingly, it has been studied widely for different languages \cite{prabhu2019kannada,jangid2018handwritten,el2016cnn}. However, this is not the case for multi-script digit recognition works that encourage the development of robust and multipurpose systems. Whereas in practice it is possible to see multiple scripts in a document. More importantly working on multi-script recognition opens a way for multi-task learning (MTL), considering the script classification as an auxiliary task for instance. Deep learning methods proved to show a very good recognition performance on such classification tasks. Apart from the success stories of recognition performance, the requirement for large amount of data, the issue of over-fitting, and computation cost of the complex models has remained a challenge in the area of deep learning. On the other hand the introduction of multi task learning seems to have resolutions for  that. With multi-task learning one can address multiple problems reducing the requirement of having individual models\cite{zhang2017survey}. On top of this, it has shown to be good at regularizing the models which prevent from over-fitting\cite{sener2018multi}. One can also use advantage of multi-task learning to increase amount of dataset which is usually required in machine learning. However, multi-task learning by itself is not free from challenges. Combining the losses of the different tasks, tuning the hyper-parameters, and using the estimate  of one task as a feature to another task are the major challenges in multi-task learning\cite{kendall2018multi}.

\begin{figure}
\centering
\begin{minipage}{.4\textwidth}
	\centering
	\includegraphics[width=\textwidth]{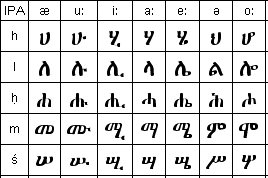}
	\caption{Parts of Amharic alphabet.} \label{amh}
\end{minipage}\qquad
\begin{minipage}{.5\textwidth}
	\centering
	\includegraphics[width=0.8\textwidth]{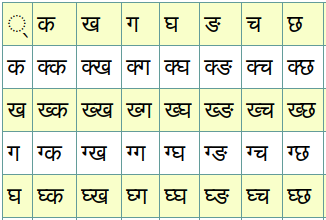}
	\caption{Parts of Devanagari alphabet.} \label{dev}
\end{minipage}
\begin{minipage}{.7\textwidth}
\vspace{5mm}
	\centering
	\includegraphics[width=0.6\textwidth]{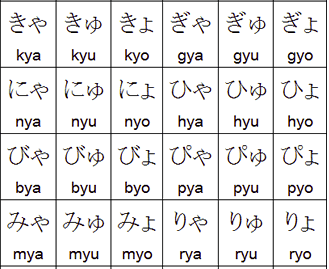}
	\caption{Parts of Japanese Hiragana alphabet.} \label{jap}
\end{minipage}%
\end{figure}

In this study we make use of multi-task learning and also avoid one of the challenges which is combining the different weighted losses. First we will introduce the formulation of a multi-task learning setting from the individual tasks. The motivation behind this formulation is to bring the problem of Amharic, Indian, Japanese, and related character recognition \cite{gondere2019handwritten,jangid2018handwritten,alom2018handwritten,tsai2016recognizing} to a more general setting so that researchers contribute to the solution with ease. In these languages the alphabets can be organized in a matrix form where one can exploit the information available over the rows and columns as they exhibit similarities, Fig.~\ref{amh},\ref{dev},\ref{jap}. Since there is no baseline with this method, we aim at presenting an exploratory investigation towards a higher a goal. Hence, in this study we organize the main task (classifying the exact label) in to additional rows and columns of different classification tasks as shown in Table~\ref{tab1}. All these digits are Hindu–Arabic numeral systems where the widespread Western Arabic numerals are used with Latin scripts whereas the Eastern Arabic numerals are used with Arabic scripts. However, Kannada with its own script is the official and administrative language of the state of Karnataka in India \cite{ashiquzzaman2017handwritten,prabhu2019kannada}. In this study, this general method will also be applied to the specific case of Amharic handwritten character recognition.  

Finaly we will compare three models. The first one is a baseline model to classify each label as a thirty class \((3 scripts \times 10 digits)\) classification problem. Using a related task as an auxiliary task for MTL is the classical choice\cite{ruder2017overview}. Hence, the second model employees a conventional multi-task learning considering the classification of the rows (scripts) and columns (digits) as auxiliary tasks. The third one which is the proposed model also applies multi-task learning however with a new way of controling and exploiting the information contained in the related tasks. This is basically done by creating a four class classification problem as an auxilary task. The four class lables indicate whether the main task is good at identifying the digit, the language, the lable (both digit and language), or none. By doing this we will get the information regarding the training behaviour of the different tasks which can be used to help and control the main task. For this study since we gave emphasis to show the useful formulation and advantage of multi-task learning, we adapted the ResNet\cite{he2016deep} pretrained model to build our models. The rest of this paper is organized as follows: related works are reviewed in the next section. Section 3 outlines the methodology followed for the study and experimental results are discussed under Section 4. Finally, conclusion and future works are forwarded. 

\subsubsection{Contributions of the Paper:} 
\renewcommand{\theenumi}{\roman{enumi}}
\begin{enumerate}
  \item Presents the possible formulation of individual tasks in to multi-task learning setting
  \item Proposes a novel way of exploiting axillary tasks to regularize and help the main task
  \item Demonstrates the proposed method on the specific case of Amharic handwritten character recognition
\end{enumerate}

\begin{table}[ht]
\caption{Organization of the individual tasks in to multi-task settings.}\label{tab1}
\begin{tabular}{l}
\includegraphics[width=\textwidth]{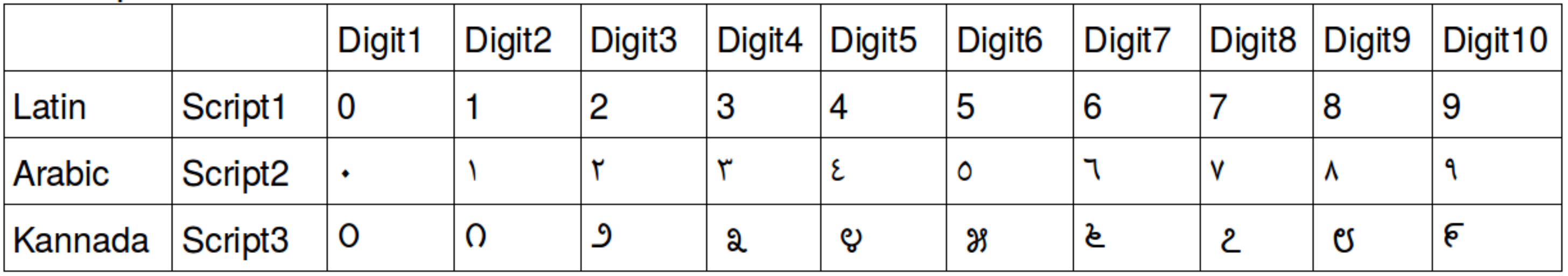}
\end{tabular}
\end{table}

\section{Related Works}
There are some related works on multi-script recognition and a few of them employ multi-task learning. Sadeghi et al.\cite{sadeghi2017bilingualism} performed a comparative study between a monolingual training and bilingual training (Persian and Latin digits) using deep neural networks. They have reported the superior performance of bilingual networks in handwritten digit recognition, thereby suggesting that mastering multiple languages might facilitate knowledge transfer across similar domains. Bai et al. \cite{bai2014image} proposed shared-hidden-layer deep convolutional neural network (SHL-CNN), the input and the hidden layers are shared across characters of different tasks while the final soft-max layer is task dependent. They have used Chinese and English superimposed texts and show that the SHL-CNN reduce recognition errors by 16-30\% relatively compared with models trained by characters of only one language. Maitra et al. \cite{maitra2015cnn} employed six databases: MNIST, Bangla numerals, Devanagari numerals, Oriya numerals, Telugu numerals and Bangla basic characters. They used the larger class (Bangla basic characters) pretrained on CNN as a feature extractor with aim  to show the transfer learning that result in a good performance of other scripts with smaller class. All of the above mentioned works didn’t address to balance the effects of the related tasks.

Multi-Task Learning (MTL) is a learning paradigm in machine learning and its aim is to leverage useful information contained in multiple related tasks to help improve the generalization performance of all the tasks\cite{zhang2017survey}. Technically, it is also optimizing more than one loss function in contrast to single-task learning. We can view multi-task learning as a form of inductive transfer. Inductive transfer can help improve a model by introducing an inductive bias provided by the auxiliary tasks, which cause the model to prefer hypotheses that explain more than one task \cite{zhang2017survey,ruder2017overview}. According to Zhang et al.\cite{zhang2017survey} MTL algorithms are classified into five categories: feature learning (feature transformation and feature selection approaches), low-rank, task clustering, task relation learning, and decomposition. The widely used approach of MTL including this study is homogeneous MTL and parameter based MTL with decomposition approach. In this case the tasks are decomposed with their relevance and usually the main task remains unpenalized. Zhang et al.\cite{zhang2017survey} suggest the decomposition as a good MTL approach with the limitation of the black box associated with coefficients and forward future work emphasize its formalization since there is no guaranty that MTL is better than single task.

Ruder \cite{ruder2017overview} introduced the two most common methods for MTL in Deep Learning, soft parameter sharing and hard parameter sharing. As in most computer vision tasks this study uses hard parameter sharing where the hidden layers between all tasks are shared while keeping task specific output layers. The author \cite{ruder2017overview} further stress that only a few papers have looked at developing better mechanisms for MTL in deep neural networks and our understanding of tasks, their similarity, relationship, hierarchy, and benefit for MTL is still limited. Since multitask learning models are sensitive to task weights and task weights are typically selected through extensive hyperparameter tuning, Guo et al. \cite{guo2018dynamic} introduced dynamic task prioritization for multitask learning. This avoids the imbalances in task difficulty which can lead to unnecessary emphasis on easier tasks, thus neglecting and slowing progress on difficult tasks.

Research works on Amharic document recognition in general lack combined efforts mainly due to unavailabilty of publicly available standard dataset. Accordingly different techniques are applied in different times without tracing and following a common baseline. It is worth mentioning the work done by Assabie et al.\cite{assabie2011offline} for handwritten Amharic word recognition. Betselot et al.\cite{reta2018amharic} also worked on handwritten Amharic character recognition. Both works used their own datasets and employed conventional machine learning techniques. Recently there are some encouraging works emerging on Amharic character recognition applying deep learning techniques. The different authors emphasized on different types of documents including printed \cite{addis2018printed,belay2019amharic,gebretinsae2019handwritten}, ancient \cite{negashe2020modified,demilew2019ancient}, and handwritten documents \cite{gondere2019handwritten,gebretinsae2019handwritten,abdurahman2019handwritten}. Accordingly the research efforts in this regard lack complementing each other and improving results based on a clear baseline.

\section{Methodology}
This section outlines the dataset preparation and organization of the models for the experiments.
\subsection{Dataset Preparation}
We have used the publicly available datasets MNIST \cite{romanuke2016training}, MADBase \cite{el2016cnn}, and Kannada MNIST \cite{prabhu2019kannada} for Latin, Arabic, and Kannada handwritten digit scripts respectively. All these datasets are \(28\times28\) pixel size images and they have equal size of data-sets which is 60,000 for training and 10,000 for test. We have used 16\% of the training set for validation with a balanced stratified split.

The dataset for Amharic handwritten character recognition experiment was organized from  Assabie et al. and Samuel et al. \cite{assabie2009comprehensive,gondere2019handwritten}.  It was organized to 77 characters with 11 (row) by 7 (column) tabular structure as shown in Table~\ref{tab4}. This is done intentionally to minimize the number of classes as compared to the number of samples available per character, that is 150. Another reason is to see the application of the proposed method in a balanced dataset setup on visually similar characters.

\begin{table}[ht]
\centering
\caption{77 visually similar Amharic characters.}\label{tab4}
\begin{tabular}{l}
\includegraphics[width=0.4\textwidth]{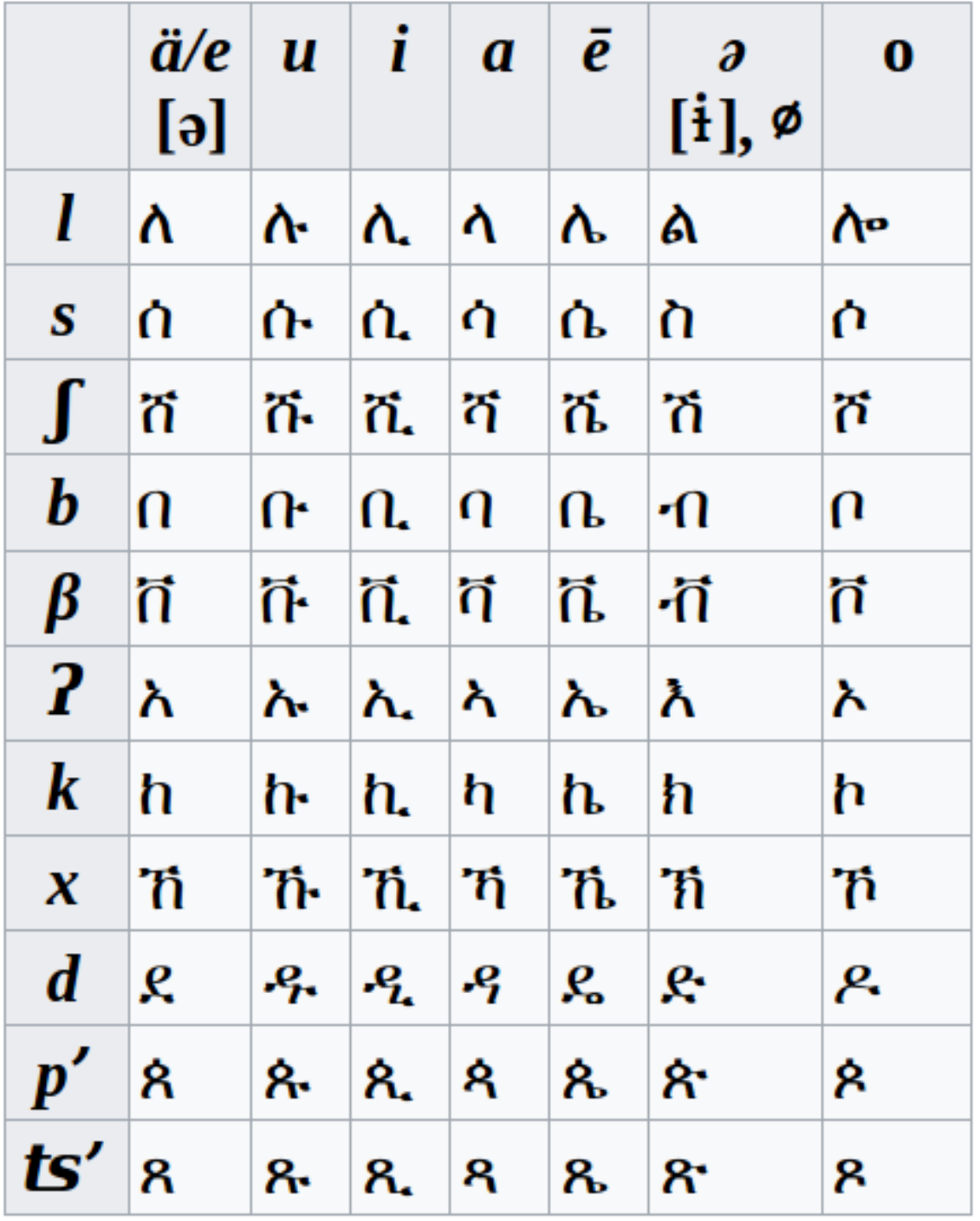}
\end{tabular}
\end{table}

\subsection{Organization of the Experiments}
\begin{figure}
\centering
\includegraphics[width=0.7\textwidth]{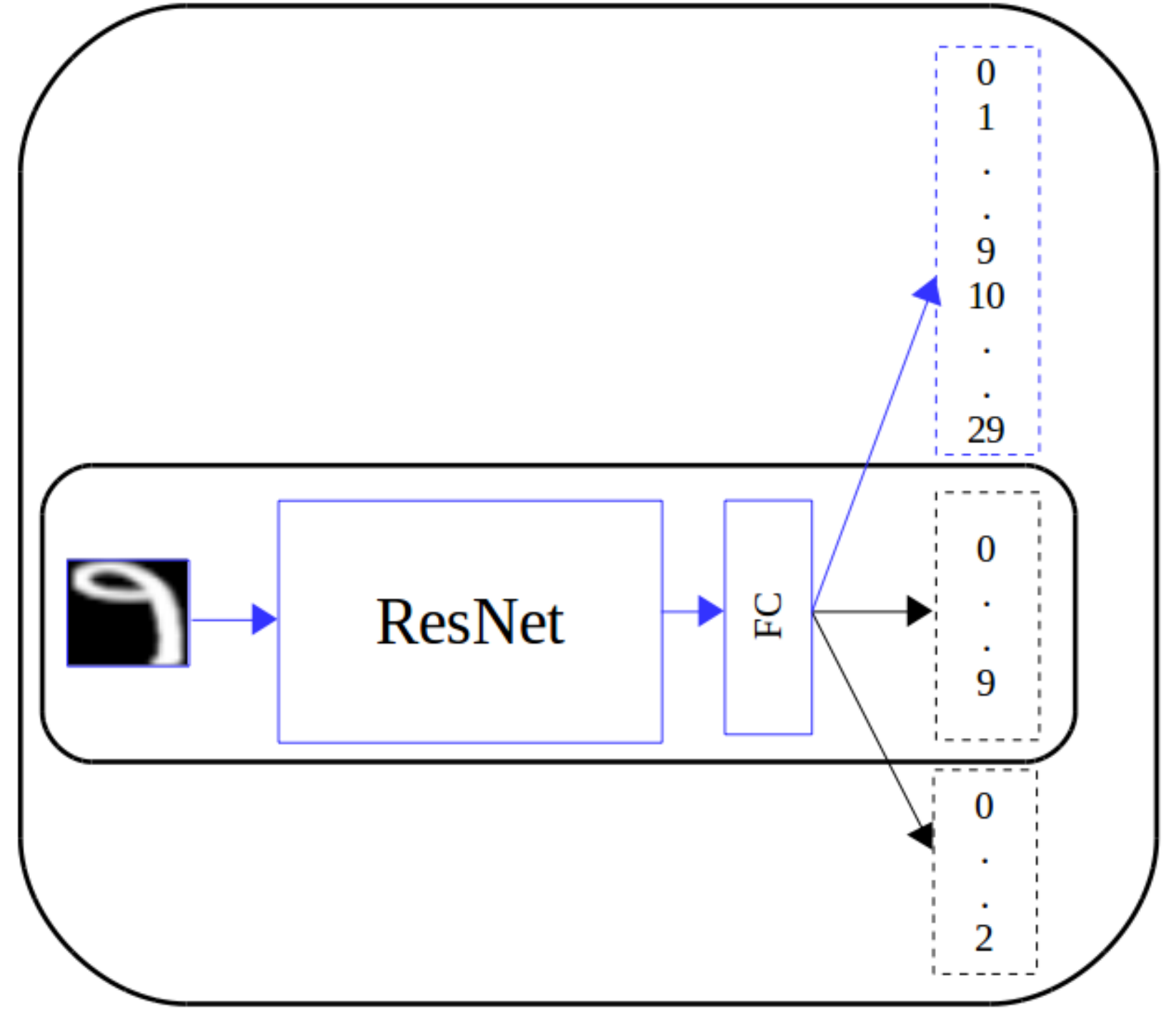}
\caption{Models Stucture. Blue line shows the baseline model, the outer rectangle shows the multi-task models, and the inner rectangle represents the individual (single task) model.} \label{fig1}
\end{figure}

In this study the main task is to classify each label from all the three scripts. That is, a trained model is expected to classify each image as it is “X digit from Y script”. Hence, our baseline model will be a thirty class \((3 scripts \times 10 digits)\) classification model, the blue line in Fig.~\ref{fig1}. Another baseline is the usual multi-task learning, the outer rectangle in Fig.~\ref{fig1}, which uses the advantages of reformulating the main problem in to other two auxiliary tasks. The third approach, proposed, will introduce a novel way of integrating multi-task learning to extract relevant information from the auxiliary tasks while balancing the effect on each other.From the usual multi-task learning we show the optimum performance by tuning the loss weights which is better than the first baseline 30 class vanilla model. For the sake of just giving a glance on how the specialized models perform, we have also experimented the three independent single task models. All the experiments conducted in this study are described in Table~\ref{tab2}.

The proposed method removes the two auxiliary tasks and introduces a one reformuated auxilary task instead. That is a four class classification problem including getting both the row and column (the label), only the row (language), only the column (digit), and missing both. This information can be obtained from the main task itself by converting the predicted label in to row and column using the formulas \( row = label \mathbf{div} 10\) and \(column = label\bmod10\) respectively. This helps to learn the properties of the characters like how they confuse the model during the training with out affecting each other. We give highest number that is \(3\) as a label for getting both rows and columns which in turn signals over fitting. We also add this numbers with in batches to use as a factor to be multiplied to control the loss of the main task. That is the more we know this label the more the loss will be. Therefore, the model prefers to minimize the second loss instead. That is predicting the properties of the characters in to four classes. This balances and controls the whole training process. 

Likewise for Amharic handwritten character recognition the same procedure will be followed. Here the row will be \(11\) instead of \(3\) and the column will be \(7\) instead of \(10\). The \(3 \times 10 = 30\) class classification problem will now be \(11 \times 7 = 77\) class classification problem in the case of Amharic characters.  

\begin{minipage}{\textwidth}
\vspace{5mm}
\begin{equation}
L^{Base}=l(y, \hat{y})
\label{eq1}
\end{equation}
\begin{equation}
L^{Wloss}=l(y, \hat{y})+\sigma_1 \cdot l(y_1, \hat{y_1})+\sigma_2 \cdot l(y_2, \hat{y_2})
\label{eq2}
\end{equation}
\begin{equation}
L^{New}=factor \cdot l(y, \hat{y})+l(y_a, \hat{y_a})
\label{eq3}
\end{equation}
\( \textrm{where in all the equations } y, y_1, y_2, \hat{y}, \hat{y_1}, \hat{y_2} \textrm{ are the ground truth and } \\ \textrm{prediction of their respective label, digit, and script classes and } \\ l \textrm{ is Cross Entropy Loss.}\)
\end{minipage}

\begin{table}
\caption{Description of the different models in the experiment.}\label{tab2}
\begin{tabular}{|l|c|p{7.5cm}|}
\hline
Model Name &  Equation & Description\\
\hline
Lat &  (1) & Single task model trained on the Latin digits\\

Arab &  (1) & Single task model trained on the Arabic digits\\

Kan &  (1) & Single task model trained on the Kannada digits\\

Base &  (1) & Single task model trained on all the three scripts\\

Wloss &  (2) & Multi-task model with weighted loss of 0.2,0.65 for sigma 1 and 0.3,0.35 for sigma 2 for multi-script and Amharic recognition respectively\\

New &  (3) & The newly proposed model\\
\hline
\end{tabular}
\end{table}

\section{Experimental Results}
Due to our emphasis to show the useful formulation and advantage of multi-task learning over individual tasks, in all the models in these experiments we have adapted the ResNet pretrained model from torchvision.models. We have used a mini batch size of 32 and Adam optimizer. All these configurations are kept unchanged between the individual, baseline, and the multi-task models. All the experiments were performed using Pytorch 1.3.0 machine learning framework on GPU nodes connected to computing cluster at Information Systems and Machine Learning Lab (ISMLL), University of Hildesheim.

\begin{figure}
\includegraphics[width=\textwidth]{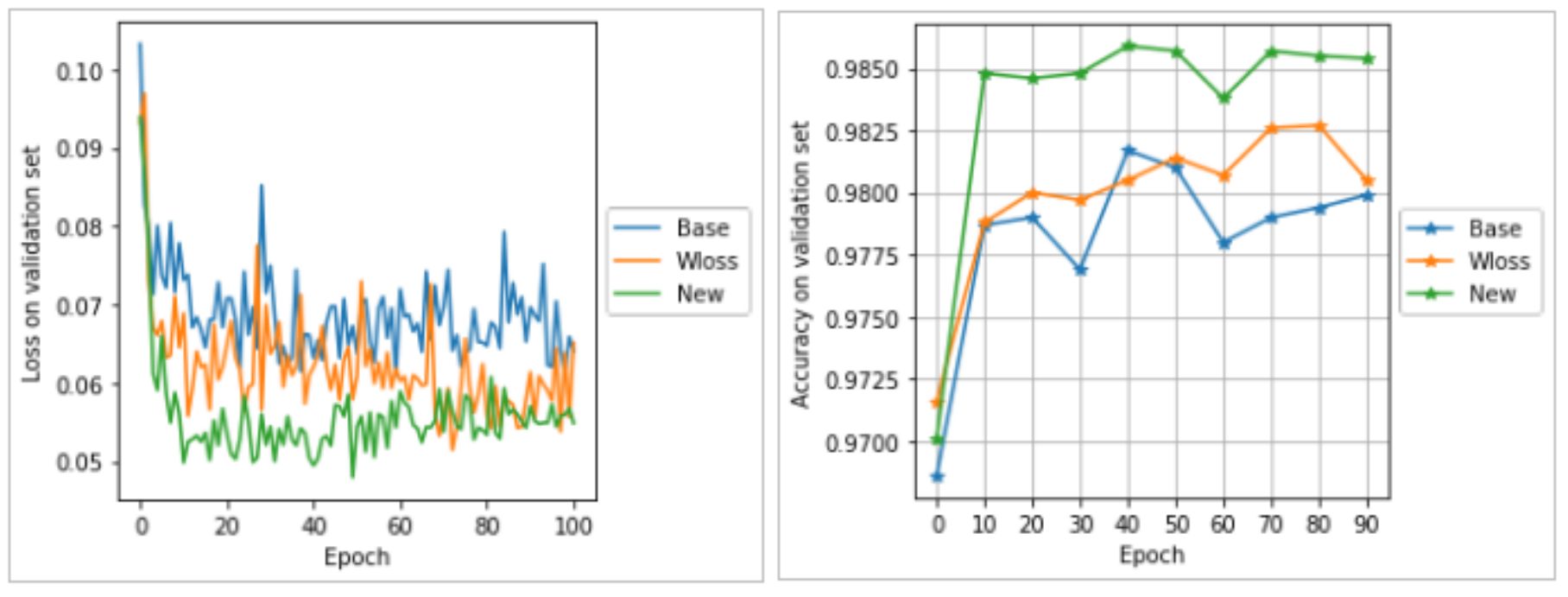}
\caption{The learning behavior of the models, multi-script.} \label{fig2}
\end{figure}

\begin{figure}
\includegraphics[width=\textwidth]{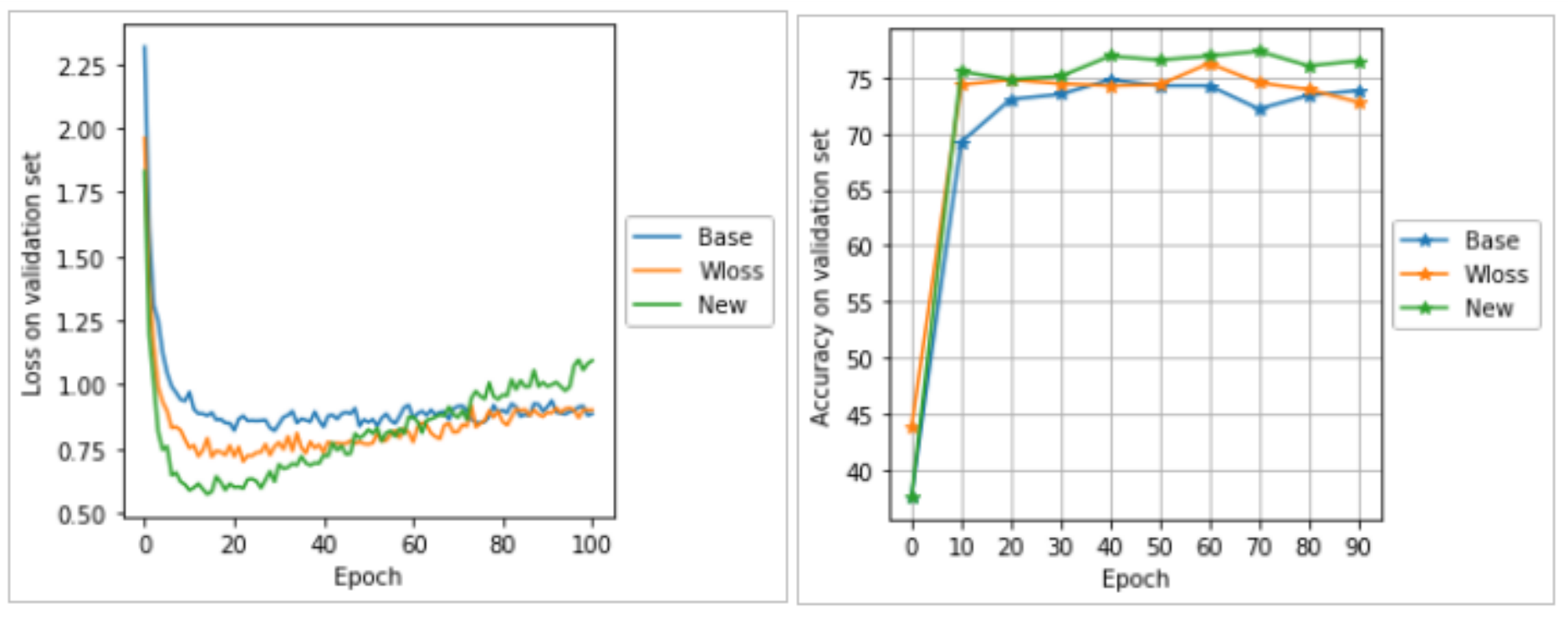}
\caption{The learning behavior of the models, Amharic.} \label{fig7}
\end{figure}

\begin{figure}
\includegraphics[width=\textwidth]{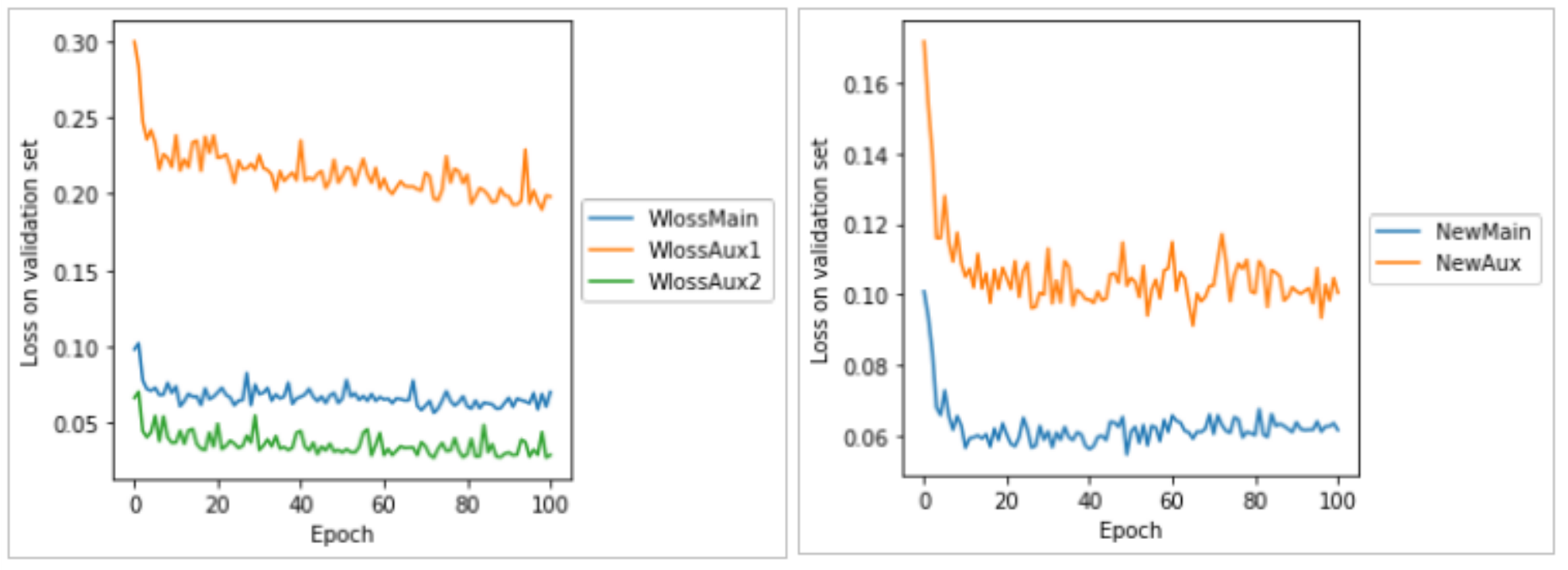}
\caption{Regularizing the main task, multi-script.} \label{fig3}
\end{figure}

Each model run up to 100 epochs three times. The average result on test sets from the three evaluations by each model are presented in Table~\ref{tab3}. The accuracy and loss curves of the four competing models (baseline, conventional multi-task, and proposed multi-task model) are shown in Fig.~\ref{fig2} for multi-script recognition and in Fig.~\ref{fig7} for Amharic recognition.  Further Fig.~\ref{fig3} show how the proposed multi-task model regularizes the main task as compared to the conventional multi-task learning.

\begin{table}
\caption{Accuracy score of the models on test sets}\label{tab3}
\centering
\begin{tabular}{|l|c|c|c|c|c|c|}
\hline
Model & Latin digits & Arabic digits & Kannada digits & Average & Range & Amharic Characters\\
\hline
Lat & 98.45 & 0 & 0 & - & - & -\\

Arab & 0 & 98.49 & 0 & - & - & -\\

Kan & 0 & 0 & 96.25 & - & - & -\\

Base & 97.19 & 95.51 & 94.90 & 95.87 & 2.99 & 73.83\\

Wloss & 97.18 & 97.94 & 96.13 & 97.08 & 1.81 & 74.68\\

New & 97.85 & 98.07 & 97.23 & \textbf{97.71} & \textbf{0.84} & \textbf{75.91}\\
\hline
\end{tabular}
\end{table}

The results from Table~\ref{tab3} show the advantages gained from multi task learning. However, it is expensive to find the optimum sigmas for weighting the different losses in regard to conventional multi-task setting. Whereas the proposed multi-task approach performed best and is shown to be robust enough not to be affected by the auxiliary tasks without a need for these hidden coefficients. This is also likely to be one reason for we see the minimum range between the scores  of the proposed model.

As it can be seen in Fig.~\ref{fig3}, the proposed model enforces regularization effect for the main task. This can be seen from the oscillating behavior of the auxilary task  while maintaining a relatively smooth curve for the main task. This is an interesting behavior expected since we aim at being good on the main task. The technique incorporates the auxiliary task, their combined contribution, and the usual main task. It is formulated in a such a way that optimizing the loss implicitly observes the main task and enforces the contribution of the auxiliary task agree to the main task. Even though this tie-up is feasible for this particular problem, it is still possible to untie by introducing desired operations that allow the focus on the auxiliary tasks as well when needed. More importantly, this technique from the proposed model can open the oportunity to exploit the learned parameters of the auxiliary task during model evaluation as well. This is not common in the conventional multi-task learning where the parameters of the auxiliary tasks are wasted.  

\section{Conclusion}
This study shows a formulation of multi-task learning setting from individual tasks which can be adapted to solve related problems that can be organized in a matrix way. Therefore, the study addressed multi-script handwritten digit recognition using multi-task learning. Apart from exploiting the auxiliary tasks for the main task, this study presented a novel way of using the individual tasks predictions to help classification performance and regularize the different loss for the purpose of the main task. This finding has outperformed the baseline and the conventional multi-task learning models while avoiding weighted losses which is one of the challenges in multi-task learning. In this paper the proposed method worked for a specific case of Amharic handwritten character recognition. Hence, similar approches can also be followed to address similarly structured languages. 

Finally, we forward future works address similar multi-script multi-task learning problems encouraging the development of robust and multi-purpose systems. The generalization of the proposed model to any type of multi-task settings could also be a good future work to look. 

%
%
%

\begin{thebibliography}{8}
\bibitem{abdurahman2019handwritten}
Abdurahman, F., 2019. Handwritten Amharic Character Recognition System Using Convolutional Neural Networks. Engineering Sciences, 14(2), pp.71-87.

\bibitem{addis2018printed}
Addis, D., Liu, C.M. and Ta, V.D., 2018, June. Printed ethiopic script recognition by using lstm networks. In 2018 International Conference on System Science and Engineering (ICSSE) (pp. 1-6). IEEE.

\bibitem{alani2017arabic}
Alani, A., 2017. Arabic handwritten digit recognition based on restricted boltzmann machine and convolutional neural networks. Information, 8(4), p.142.

\bibitem{alom2018handwritten}
Alom, Z., Sidike, P., Hasan, M., Taha, T.M., and Asari1, V.K., 2018. Handwritten Bangla Character Recognition Using the State-of-the-Art Deep Convolutional Neural Networks. Computational Intelligence and Neuroscience.

\bibitem{assabie2009comprehensive}
Assabie, Y. and Bigun, J., 2009. A comprehensive Dataset for Ethiopic Handwriting Recognition. Proceedings SSBA ’09 : Symposium on Image Analysis, Halmstad University, pp 41–43

\bibitem{assabie2011offline}
Assabie, Y. and Bigun, J., 2011. Offline handwritten Amharic word recognition. Pattern Recognition Letters, 32(8), pp.1089-1099.

\bibitem{ashiquzzaman2017handwritten}
Ashiquzzaman, A. and Tushar, A.K., 2017, February. Handwritten Arabic numeral recognition using deep learning neural networks. In 2017 IEEE International Conference on Imaging, Vision and Pattern Recognition (icIVPR) (pp. 1-4). IEEE.

\bibitem{bai2014image}
Bai, J., Chen, Z., Feng, B. and Xu, B., 2014, October. Image character recognition using deep convolutional neural network learned from different languages. In 2014 IEEE International Conference on Image Processing (ICIP) (pp. 2560-2564). IEEE.

\bibitem{belay2019amharic}
Belay, B.H., Habtegebirial, T., Liwicki, M., Belay, G. and Stricker, D., 2019, September. Amharic text image recognition: Database, algorithm, and analysis. In 2019 International Conference on Document Analysis and Recognition (ICDAR) (pp. 1268-1273). IEEE.

\bibitem{das2014survey}
Das, S. and Banerjee, S., 2014. Survey of Pattern Recognition Approaches in Japanese Character Recognition. International Journal of Computer Science and Information Technologies, 5(1), pp.93-99.

\bibitem{demilew2019ancient}
Demilew, F.A., 2019. Ancient Geez Script Recognition Using Deep Convolutional Neural Network (Doctoral dissertation, Near East University).

\bibitem{el2016cnn}
El-Sawy, A., Hazem, E.B. and Loey, M., 2016, October. CNN for handwritten arabic digits recognition based on LeNet-5. In International Conference on Advanced Intelligent Systems and Informatics (pp. 566-575). Springer, Cham.

\bibitem{gebretinsae2019handwritten}
Gebretinsae Beyene, E., 2019. Handwritten and Machine printed OCR for Geez Numbers Using Artificial Neural Network. arXiv e-prints, pp.arXiv-1911.

\bibitem{gondere2019handwritten}
Gondere, M.S., Schmidt-Thieme, L., Boltena, A.S. and Jomaa, H.S., 2019. Handwritten amharic character recognition using a convolutional neural network. arXiv preprint arXiv:1909.12943.

\bibitem{guo2018dynamic}
Guo, M., Haque, A., Huang, D.A., Yeung, S. and Fei-Fei, L., 2018. Dynamic task prioritization for multitask learning. In Proceedings of the European Conference on Computer Vision (ECCV) (pp. 270-287).

\bibitem{he2016deep}
He, K., Zhang, X., Ren, S. and Sun, J., 2016. Deep residual learning for image recognition. In Proceedings of the IEEE conference on computer vision and pattern recognition (pp. 770-778).

\bibitem{jangid2018handwritten}
Jangid, M. and Srivastava, S., 2018. Handwritten devanagari character recognition using layer-wise training of deep convolutional neural networks and adaptive gradient methods. Journal of Imaging, 4(2), p.41.

\bibitem{kendall2018multi}
Kendall, A., Gal, Y. and Cipolla, R., 2018. Multi-task learning using uncertainty to weigh losses for scene geometry and semantics. In Proceedings of the IEEE Conference on Computer Vision and Pattern Recognition (pp. 7482-7491).

\bibitem{maitra2015cnn}
Maitra, D.S., Bhattacharya, U. and Parui, S.K., 2015, August. CNN based common approach to handwritten character recognition of multiple scripts. In 2015 13th International Conference on Document Analysis and Recognition (ICDAR) (pp. 1021-1025). IEEE.

\bibitem{negashe2020modified}
Negashe, G. and Mamuye, A., 2020. Modified Segmentation Algorithm for Recognition of Older Geez Scripts Written on Vellum. arXiv preprint arXiv:2006.00465.

\bibitem{prabhu2019kannada}
Prabhu, V.U., 2019. Kannada-mnist: A new handwritten digits dataset for the kannada language. arXiv preprint arXiv:1908.01242.

\bibitem{rajput2010printed}
Rajput, G.G., Horakeri, R. and Chandrakant, S., 2010. Printed and handwritten kannada numeral recognition using crack codes and fourier descriptors plate. International Journal of Computer Application (IJCA) on Recent Trends in Image Processing and Pattern Recognition (RTIPPR), pp.53-58.

\bibitem{reta2018amharic}
Reta, B.Y., Rana, D. and Bhalerao, G.V., 2018, May. Amharic handwritten character recognition using combined features and support vector machine. In 2018 2nd International Conference on Trends in Electronics and Informatics (ICOEI) (pp. 265-270). IEEE.

\bibitem{romanuke2016training}
Romanuke, V.V., 2016. Training data expansion and boosting of convolutional neural networks for reducing the MNIST dataset error rate.

\bibitem{ruder2017overview}
Ruder, S., 2017. An overview of multi-task learning in deep neural networks. arXiv preprint arXiv:1706.05098.

\bibitem{sadeghi2017bilingualism}
Sadeghi, Z., Testolin, A. and Zorzi, M., 2017, October. Bilingualism advantage in handwritten character recognition: A deep learning investigation on Persian and Latin scripts. In 2017 7th International Conference on Computer and Knowledge Engineering (ICCKE) (pp. 27-32). IEEE.

\bibitem{sener2018multi}
Sener, O. and Koltun, V., 2018. Multi-task learning as multi-objective optimization. In Advances in Neural Information Processing Systems (pp. 527-538).

\bibitem{tsai2016recognizing}
Tsai, C., 2016. Recognizing handwritten Japanese characters using deep convolutional neural networks. university of Stanford in Stanford, California.

\bibitem{zhang2017survey}
Zhang, Y. and Yang, Q., 2017. A survey on multi-task learning. arXiv preprint arXiv:1707.08114.
\end{thebibliography}
%

\end{document}